\pdfoutput=1

\documentclass[11pt]{article}

\usepackage{EMNLP2023}

\usepackage{times}
\usepackage{latexsym}

\usepackage{graphicx}
\usepackage{enumitem}
\usepackage[namelimits]{amsmath} 
\usepackage{amssymb}             
\usepackage{amsfonts}            
\usepackage{mathrsfs}            

\usepackage{caption,subcaption}
\usepackage{multirow}

\usepackage{booktabs} 

\usepackage[T1]{fontenc}

\usepackage[utf8]{inputenc}

\usepackage{microtype}

\usepackage{inconsolata}

\title{CPRM: A LLM-based Continual Pre-training Framework for Relevance Modeling in Commercial Search}

\author{Kaixin Wu\textsuperscript{1}, Yixin Ji\textsuperscript{1}, Zeyuan Chen\textsuperscript{1}, Qiang Wang\textsuperscript{2}, Cunxiang Wang\textsuperscript{3}\\ 
\textbf{Hong Liu\textsuperscript{1}, Baijun Ji\textsuperscript{4}, Jia Xu\textsuperscript{1}\thanks{\quad Corresponding author.}, Zhongyi Liu\textsuperscript{1}, Jinjie Gu\textsuperscript{1}, Yuan Zhou\textsuperscript{1}, Linjian Mo\textsuperscript{1}} \\
\textsuperscript{1}{Ant Group}, \textsuperscript{2}{Hithink RoyalFlush AI Research
Institute} \\
\textsuperscript{3}{Westlake University}, \textsuperscript{4}{Soochow University} \\
\texttt{daniel.wkx@antgroup.com}}

\begin{document}
\maketitle
\begin{abstract}
Relevance modeling between queries and items stands as a pivotal component in commercial search engines, directly affecting the user experience. Given the remarkable achievements of large language models (LLMs) in various natural language processing (NLP) tasks, LLM-based relevance modeling is gradually being adopted within industrial search systems. Nevertheless, foundational LLMs lack domain-specific knowledge and do not fully exploit the potential of in-context learning. Furthermore, structured item text remains underutilized, and there is a shortage in the supply of corresponding queries and background knowledge. We thereby propose \textbf{CPRM} (\textbf{C}ontinual \textbf{P}re-training for \textbf{R}elevance \textbf{M}odeling), a framework designed for the continual pre-training of LLMs to address these issues. Our CPRM framework includes three modules: 1) employing both queries and multi-field item to jointly pre-train for enhancing domain knowledge, 2) applying in-context pre-training, a novel approach where LLMs are pre-trained on a sequence of related queries or items, and 3) conducting reading comprehension on items to produce associated domain knowledge and background information (e.g., generating summaries and corresponding queries) to further strengthen LLMs. Results on offline experiments and online A/B testing demonstrate that our model achieves convincing performance compared to strong baselines. 
\end{abstract}

\section{Introduction}
Relevance modeling is designed to evaluate the correlation between queries and items, an essential component of commercial search engines and crucial for the user experience. Mini-app service search is a common search application scenario. Unlike traditional e-commerce searches that only provide product search functions, mini-app services encompasses numerous scenarios such as livelihoods, government affairs, transportation, healthcare and dining. Moreover, the item consists of structured data with multiple fields; for instance, a hospital mini-app might include fields like title, keywords, category and description. Considering the diverse scenes and the complexity of structured data across multiple fields, conducting relevance modeling within the such search scenario poses a significant challenge. The current relevance model in commercial search systems is a semantic matching model, leveraging LLMs combined with domain-annotated data through supervised fine-tuning (SFT) methods. These LLMs like GPT-3~\citep{brown2020language}, GLM~\citep{du2022glm}, LLaMA~\citep{touvron2023llama1}, Qwen~\citep{bai2023qwentechnicalreport,yang2024qwen2technicalreport}, having more extensive parameters and utilizing a massive corpora of texts during training compared to previous pre-trained models such as BERT~\citep{Devlin2019BERTPO}, RoBERTa~\citep{liu2019roberta} and XLNet~\citep{yang2020xlnet}, demonstrate superior performance in semantic matching tasks. 

Despite the great success of LLMs, there still remain certain limitations in their application to relevance modeling. Firstly, LLMs are pre-trained on a broad range of data sources~\citep{brown2020language,du2022glm,touvron2023llama1,touvron2023llama2}, which do not afford special attention to particular domains~\citep{wu2023bloomberggpt,cui2023chatlaw,xiong2023doctorglm}, resulting in a lack of domain-specific knowledge. Besides, queries tend to be colloquial and present in short-text form, whereas items are typically expressed in a more formal long-text form, leading to a ``semantic gap''~\citep{lian2019endtoend,qi2020prophetnet,kumar2021neural} between their representations. Secondly, the pre-training phase of LLMs is ``task-agnostic''~\citep{brown2020language}, which impedes direct connection with downstream tasks and precludes the possibility of in-context pre-training enhancements tailored for these tasks~\citep{min2022metaicl,gu2023pretraining}. Finally, the item tends to be highly structured and difficult to leverage, which prevents LLMs from fully realizing their potential with such data.

To address the above problems, we investigate a \textbf{C}ontinual \textbf{P}re-training approach of LLMs for \textbf{R}elevance \textbf{M}odeling, CPRM for short. Initially, we introduce a pre-training technique using pairs of queries and multi-field item as inputs. This method enables the LLMs to explicitly model the semantic representations between queries and items, thus bridging the semantic gap between them. Subsequently, we collect sets of semantically similar queries and items based on user click logs, then further refine these sets through semantic modeling to filter out semantically irrelevant cases. Following that, we reorder these refined sets according to semantic similarity and ultimately construct in-context pre-training instances via prompting techniques. Utilizing this approach to data construction, LLMs are able to make better predictions within such contexts during the training process, which benefits the efficient learning current domain knowledge for LLMs. Lastly, we employ an LLM with larger parameters (called teacher LLM)  to conduct reading comprehension on structured item data, facilitating the generation of relevant domain knowledge for pre-training, which can be considered as a secondary development and exploitation of the item. More specifically, we leverage teacher LLM to summarize and paraphrase item to produce fluent domain knowledge, while also guiding teacher LLM to produce background information related to the items. Additionally, we prompt teacher LLM to create diverse queries and provide further reasons for their generation. In summary, the contributions of this paper are as follows: 
\begin{itemize} [leftmargin=*]
    \item To our knowledge, we are the first to systematically propose a continual pre-training approach of LLMs specifically designed for relevance modeling tasks. 
    
    \item We propose a CPRM framework with three components. Firstly, the joint pre-training of queries and multi-field item to enhance domain knowledge of LLMs. Secondly, in-context pre-training by constructing collections of semantically similar queries or items. And thirdly, reading comprehension of structured items employed to strengthen the capabilities of LLMs further.
    
    \item Our approach has been validated on real-world industry data, outperforming strong baselines significantly in both offline experiments and online A/B testing. 
\end{itemize}
\begin{figure}[!t]
    \centering
    \includegraphics[width=1.0\linewidth]{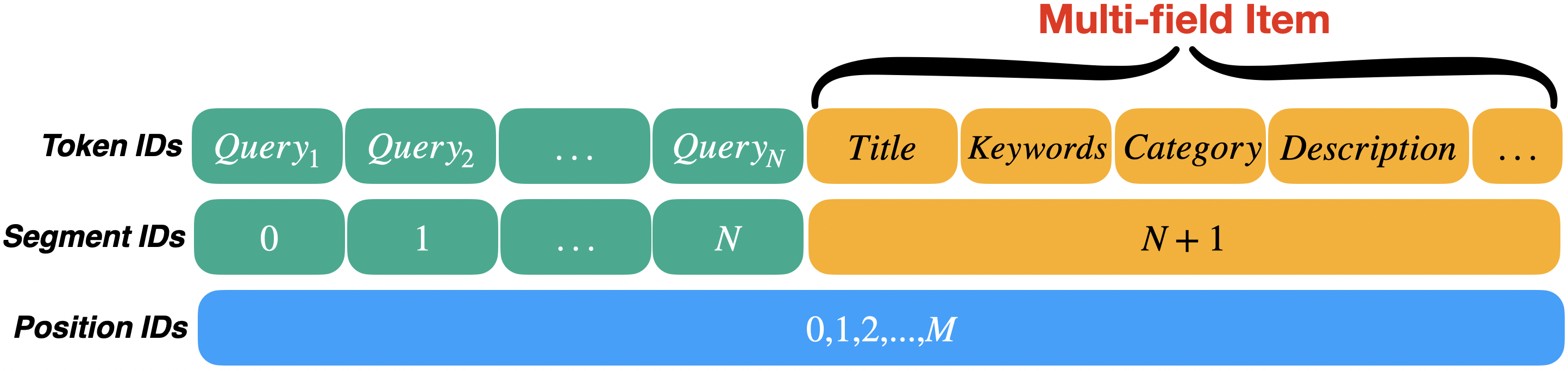}
    \caption{Joint queries and multi-field item for pre-training. An example of the mini-app search scenario. }
    \label{fig:DKE}
\end{figure}
\section{Related Work}
Relevance modeling in search corresponds to the semantic matching task in NLP. With the advancement of neural network and pre-trained models, deep semantic matching models have become mainstream. 
Deep semantic matching models are categorized into two types: representation-based ~\citep{shen2014learning,palangi2015semantic,rao2019bridging} and interaction-based methods~\citep{chen2016enhanced,hu2014convolutional,pang2016text,parikh2016decomposable}. The former focuses on learning low-dimensional representations, while the latter emphasizes capturing the interactions between inputs. The representation-based model with independently encoded inputs struggles to capture complex correlations, whereas interaction-based methods that concatenate the two inputs for semantic computation can alleviate this issue.

In recent years, pre-trained models like BERT~\citep{Devlin2019BERTPO} has show its superiority on natural language understanding (NLU) tasks. Consequently, both representation-based and interaction-based methods have begun leveraging the capabilities of these pre-trained models for semantic modeling. 
Most recently, LLMs like GPT-3~\citep{brown2020language}, GLM~\citep{du2022glm}, LLaMA~\citep{touvron2023llama1}, Qwen~\citep{bai2023qwentechnicalreport,yang2024qwen2technicalreport} pre-trained extensive volumes of data with numerous parameters have garnered significant performance in language understanding, generation and reasoning tasks. Compared to traditional pre-trained models like BERT~\citep{Devlin2019BERTPO}, LLMs possess significant advantages in both the scale of pre-training data and the quantity of model parameters, leading to their evident superiority in performance across a variety of downstream tasks. Recent research work~\citep{sun2023chatgpt,spatharioti2023comparing,zhu2024large} indicates that combining LLMs with downstream applications presents significant potential, LLMs can achieve competitive or even superior results compared to traditional supervised methods on information retrieval benchmarks. Some research leverage LLMs for relevance modeling in search engines, adopting approaches such as behavior-augmented~\citep{chen2023beyond,chen2024boostingllmsdrivenrelevancemodeling} or robust learning~\citep{liu2024boosting} to improve the capability of relevance modeling. Our work mainly focuses on enhancing LLMs from the perspective of continual pre-training for relevance modeling.
LLMs are pre-trained on a wide variety of data sources~\citep{brown2020language,du2022glm,touvron2023llama1,touvron2023llama2} without pay more attention on specific domains, resulting in a lack of domain knowledge. On the other hand, the pre-training phase of LLMs is task-agnostic~\citep{brown2020language}, making it difficult to direct connect with downstream tasks. This also means we can't easily customize the pre-training process to better fit those tasks~\citep{min2022metaicl,gu2023pretraining}. 
Previous work of injecting domain knowledge involves continued training of pre-trained models on domain-specific data~\citep{gururangan2020dont,shi2023rethinking},  as well as incorporating knowledge graphs~\citep{liu2019kbert} or selectively masking important information~\citep{gu2020train,xu2023kilm,sanyal2023apollo,zhou2023commonsense}.
Another line of research aims to enhance the pre-training for downstream tasks, which simple concatenate relevant documents together for in-context pre-training~\citep{min2022metaicl,gu2023pretraining,shi2024incontext}. However, these approaches assume that downstream tasks contain only a single domain representation, neglecting the possibility of there being multiple or more. For instance, in commercial search relevance tasks, queries and items belong to two distinct domains with substantial differences. Our research work involves injecting domain knowledge and conducting in-context pre-training simultaneously, while being able to establish a connection between the two domains.
\section{Problem Formulation}
Given a query $Q$ and an item $I$, LLM-based relevance modeling in search engines aims to predict the relevance degree between them. Essentially, referring
to PET~\citep{schick2021exploitingclozequestionsshot}, we first design the prompt $\textrm{P}(Q, I)$, and LLM determines which verbalizer $v$ (e.g., “no” or “yes”) is the most likely candidate for “$[\textrm{Mask}]$” conditioned on the likelihood $\textrm{M}(v|\textrm{P}(Q, I))$. The above process is defined as follows:
\begin{align}
& \textrm{P}(Q, I) = Is \ [Q] \ and \ [I] \ related? \ \ [\textrm{Mask}] \\
& y = \textrm{M}(v \ | \ \textrm{P}(Q, I)), \quad \textrm{for} \ v \in \{\textrm{no, yes}\},
\end{align}
where relevance label $y \in \{0, 1\}$ can be associated with a verbalizer (e.g., “no” or “yes”) from the vocabulary of LLMs to represent “irrelevant” and “relevant” between $Q$ and $I$ respectively. To enable the adaptation of general LLMs to the relevance modeling task, SFT operation is selected for training with the cross-entropy loss function. Consequently, the relevance degree could be given from LLMs for subsequent applications in search scenarios.
\begin{figure*}[!t]
    \centering
    \includegraphics[width=0.97\linewidth]{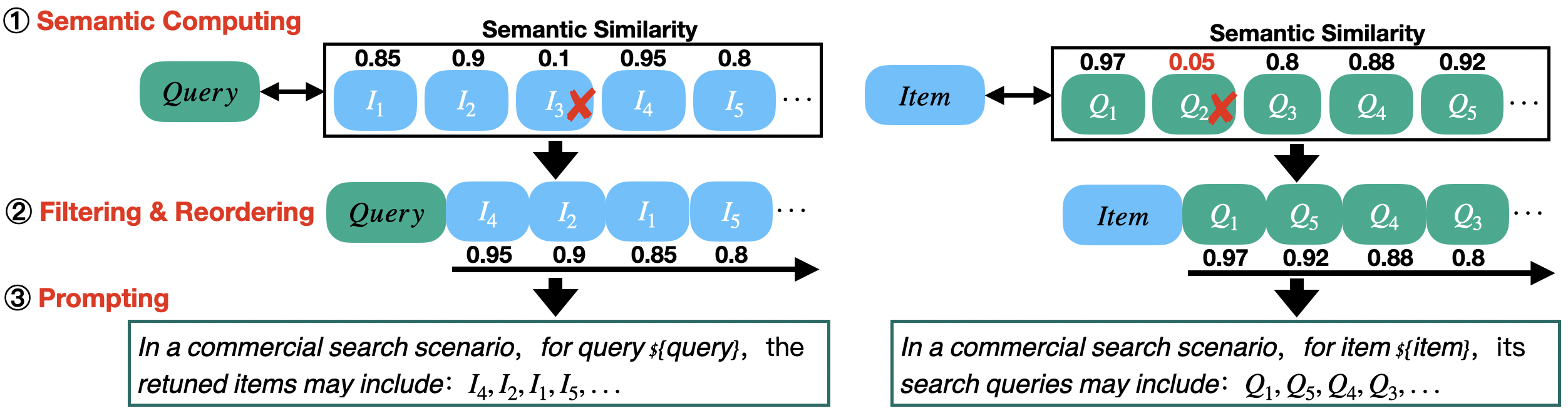}
    \caption{In-context pre-training instances construction. The left and right figures represent the ICP instances constructed from similar item sets and similar query sets respectively.}
    \label{fig:ICP}
\end{figure*}
\section{Methodology}
In this section, we present the details of the CPRM framework, including Domain Knowledge Enhancement (DKE), In-Context Pre-training (ICP) and Reading Comprehension Distillation (RCD). 
\subsection{Domain Knowledge Enhancement (DKE)}
Different from conventional pre-training methods, we jointly pre-train the structured item data with multiple queries as shown in Figure~\ref{fig:DKE}. Each item encompasses multiple fields, including title, keywords, category, description, etc., with the query being the most frequently searched top query for a given item. For each query or item, we employ segment embeddings to distinguish between different texts. 
For convenience, we add special tokens ``<|startofpiece|>'' and ``<|endofpiece|>'' between the queries and item as segment embeddings to differentiate them. 
Furthermore, queries and item are combined for position encoding, thereby allowing LLMs to explicitly model the relationships between them during pre-training process. Due to constraints on response time for online services, when calculating relevance scores between queries and items, only a limited number of item fields (such as title and keywords) are considered. Consequently, domain knowledge from other unused item fields, such as description, can be incorporated through continual pre-training. Considering that relevance modeling is a NLU task, we adopt both token-level masked language modeling (MLM)~\citep{devlin2018bert} and segment-level MLM pre-training strategies for LLMs. Therefore, the optimization objective is:
\begin{equation}
      \mathcal{L}(\theta) = \mathop{\textrm{min}}\limits_{\theta} \alpha \mathcal{L}_{\textrm{t-MLM}}(\theta) + (1 - \alpha) \mathcal{L}_{\textrm{s-MLM}}(\theta), 
\end{equation}
where $\theta$ is the parameters of the model, $\mathcal{L}_{\textrm{t-MLM}}(\theta)$ and $\mathcal{L}_{\textrm{s-MLM}}(\theta)$ represent token-level MLM loss and segment-level MLM loss respectively, we set $\alpha = 0.7$ in our experiments.
\subsection{In-Context Pre-training (ICP)}\label{sec:icp}
We construct in-context pre-training instances using historical click logs from real-world business search scenario. The overall idea is to build collections of semantically similar queries and items as pre-training data to further stimulate in-context learning capabilities of LLMs. The detailed data construction methodology is as follows: \\
\textbf{Coarse Screening.}\quad Utilizing the click logs, we establish a mapping from Queries to Items (denoted as $Q2I$) and from Items to Queries (denoted as $I2Q$), sorting them by the number of clicks in descending order. Consequently, within the $Q2I$ mapping, for a query $Query$ there is an associated set of items $I^{Query}=\{I_{1},I_{2},...,I_{N}\}$. These items can be considered as a preliminarily semantically related collection under the specific constraint of query $Query$. Vice versa for $I2Q$ mapping. 

\textbf{Fine Screening.}\quad Following described above, cases may be introduced that received clicks but are semantically unrelated. We employ \textit{Contriever}~\citep{izacard2022unsupervised}, a semantic model, to encode text into vectors, and then calculate the similarity between various text representations for semantic filtering. For set $I^{Query}$, when the following condition is met:
\begin{equation}
Sim(Query, I_{k}) < \sigma, \quad \textrm{for} \ k \in [1, N], \label{similarity}
\end{equation}
it signifies that $Query$ and $I_{k}$ are semantically unrelated and require filtering, where $Sim(\cdot)$ is similarity function, $\sigma$ is a threshold. 
\begin{figure}[!t]
    \centering
    \includegraphics[width=0.97\linewidth]{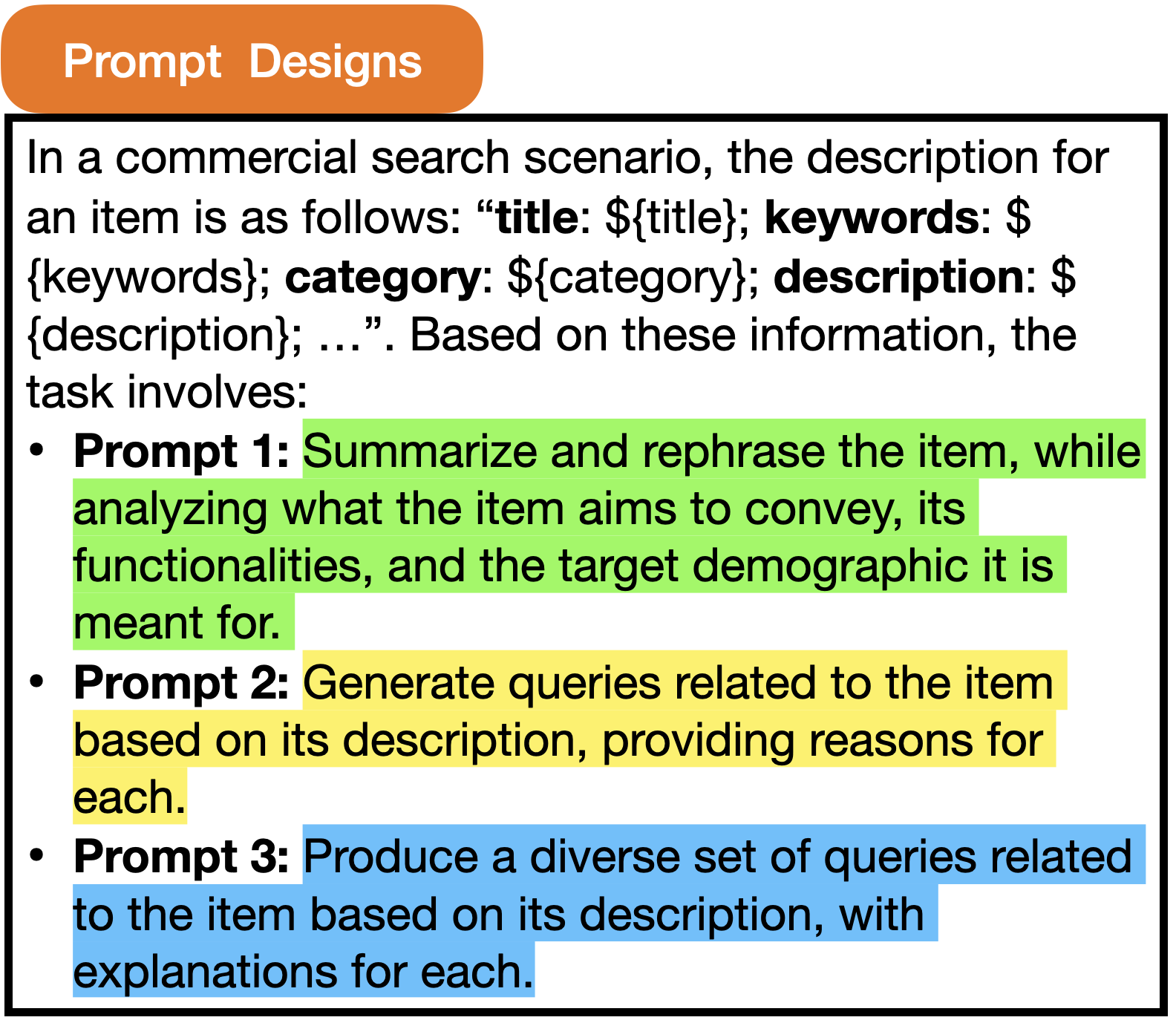} 
    \caption{Prompt for reading comprehension on item.}
    \label{fig:RCD}
\end{figure}

\textbf{Data Construction.}\quad As shown in the left of Figure ~\ref{fig:ICP}, we subsequently obtain a collection of items that are semantically relevant to the query, then sort these items by semantic similarity in ascending order. Finally, we concatenate the query with the sorted collection of items to create an ICP instance via prompting. The right of Figure~\ref{fig:ICP} illustrates how to construct ICP instances under $I2Q$ mapping, namely, obtaining a set of semantically related queries given the constraint of an item.

Why adopt this construction manner? By assembling collections of items under a specified query or collections of queries under a specified item, LLMs can make better predictions based on the context during the pre-training process, enabling more efficient learning within the current domain. Moreover, the reordering in ICP instances implicitly indicates the strength of relevance between queries and items, enabling LLMs to model the degree of their association effectively. On the other hand, by linking queries and items in our ICP instances, we enable LLMs to model their semantic representations directly.
\subsection{Reading Comprehension Distillation (RCD)}
We employ the teacher LLM for reading comprehension on items, with the prompt design shown in Figure~\ref{fig:RCD}. Assume that in a mini-app search scenario, we need to provide the mini-app's structured information like title, keywords, category, and description, and utilizing prompt template instructions to invoke teacher LLM. This generates the reading comprehension pre-training instances. 

Why design the prompt in this way? We have several reasons for this design choice. Firstly, item text is structured data and difficult to utilize, lacking in relevant background knowledge. Through summarizing and rephrasing with Prompt 1, fluent domain knowledge can be generated. Additionally, understanding and analyzing items can instruct teacher LLM in generating relevant background knowledge. Secondly, by using Prompt 2 and Prompt 3 guide teacher LLM to generate related and diversified queries, enriching the supply of suitable queries. We also instruct teacher LLM to provide further explanations for the generated queries. This approach not only facilitates the generation of relevant domain knowledge but also allows downstream models to significantly improve their understanding and handling of the item when utilizing these data. Employing teacher LLM for reading comprehension on items can be considered a secondary development and utilization of item data, enriching the domain knowledge further. Pre-training LLMs on the above data can also be seen as a process of knowledge transfer from teacher LLM to LLMs.
\section{Experiments}
\subsection{Experimental Settings}
\textbf{Dataset \& Evaluation Metrics.}\quad We utilize the real-world mini-app search scenario data for verification. The pre-training data includes three parts: DKE data (4M), ICP data (4M) and RCD data (500K). The first part is sampled from the mini-app search scenarios and consist of structured items containing multiple fields. For top 500K most frequently visited items, we sample 5 top queries based on click logs for each item, which are then concatenated with multi-field item to serve as pre-training examples for adapting relevance tasks. The second part is in-context pre-training data, where we construct these examples based on the real-world search click logs using the method described in Section~\ref{sec:icp}, and subsequently randomly sample 4M from them. The third part is reading comprehension data, for which we utilize teacher LLM (e.g. Qwen2-72B~\citep{yang2024qwen2technicalreport}) to perform reading comprehension on item data. The SFT data consists of three parts: train set (625K), valid set (35K) and test set (35K). These data are sourced from real mini-app search results and then are generated through manual annotation. The human-annotated data for relevance tasks are in format of triples <\textit{Query}, \textit{Item}, \textit{Label}>, the data statistics as shown in Table~\ref{tbl:stat}. The annotated data have only two levels of relevance: ``\textit{\#Relevant}'' and ``\textit{\#Irrelevant}''. For evaluation, we employ three widely used metrics Acc., F1 and AUC to evaluate model performance, with higher values indicating better performance. Note that AUC serves as the most important metric in relevance tasks while the others provide auxiliary supports for our analysis. 

\textbf{Implementation Details.}\quad All our pre-training experiments are conducted on the GLM-2B model. The model configuration set to 36 layers, hidden size of 2048, FFN size of 8192, and 32 attention heads. We utilize adam \citep{kingma2017adam} optimizer and the warmup steps and learning rate set 28K and $1e^{-4}$. All models are pre-trained on 8 A100 GPUs for 2 epochs and the batch size set 64. During SFT, all models are trained for 5 epochs on 8 A100 GPUs and the batch size is 8. The adam optimizer is employed and warmup steps and learning rate set to 5K and $2e^{-5}$ respectively. When constructing the ICP instances, we utilize facebook's open-source multilingual \textit{Contriever\footnote{https://github.com/facebookresearch/contriever}}~\citep{izacard2022unsupervised} model for semantic filtering. This model encodes text inputs into a 768-dimensional vector, and then calculates cosine similarity based on this vector.
\begin{table}[!t]
\centering
\resizebox{1.0\linewidth}{!}{
\begin{tabular}{cccccc}
\toprule[1.5pt]
\textbf{Dataset} & \textbf{\#Sample}  & \textbf{\#Query}  & \textbf{\#Item}  & \textbf{\#Relevant} & \textbf{\#Irrelevant}\\ \hline
 Train  & 625,292 & 92,711 & 32,219 & 370,887 & 254,405 \\
 Valid  & 35,252 & 5,016 & 8,250 & 20,023  & 15,229 \\
 Test & 35,057 & 5,426 & 8,406 & 19,164 & 15,893 \\
\bottomrule[1.5pt]
\end{tabular}
}
\caption{Data statistics (\# of numbers)}\label{tbl:stat}
\end{table}
\begin{table*}[!t]
\centering
\resizebox{0.95\linewidth}{!} {
\begin{tabular} {ll|cc|cc|cc} 
\toprule[1.5pt]
\textbf{\#} & \textbf{Model}
& \textbf{Acc.} (\%) & $\Delta_{\textrm{Acc}}$ & \textbf{F1} (\%) & $\Delta_{\textrm{F1}}$ & \textbf{AUC} (\%) & $\Delta_{\textrm{AUC}}$  \\ \hline
\hline

\multicolumn{8}{c}{\textit{Only fine-tuning on supervised datasets}} \\ \hline

1 & DSSM
& 70.32 & - & 71.21 & - & 70.64 & -  \\

2 & ReprBERT
& 80.65 & - & 82.77 & - & 80.24 & - \\

3 & BERT-Base (0.1B)
& 82.24 & - & 84.33 & - & 81.79 & - \\

4 & BERT-Large (0.3B)
& 83.47 & - & 85.65 & - & 83.01 & - \\

\hline
5 & Qwen2-0.5B
& 80.48 & - & 79.52 & - & 81.63 & -  \\

6 & Qwen2-1.5B
& 91.17 & - & 92.08 & -  & 90.90 & -  \\

\hline
7 & GLM-0.3B
& 85.93 & - & 87.32 & - & 85.64 & -  \\

8 & GLM-2B
& 91.16 & - & 91.95 & - & 91.04 & -  \\

9 & GLM-5B
& 93.53 & - & 94.12 & - & 93.41 & - \\

10 & GLM-10B
& \textbf{93.70} & - & \underline{94.26} & -  & \textbf{93.62} & -  \\

\hline \hline
\multicolumn{8}{c}{\textit{Without fine-tuning }} \\ \hline

11 & ChatGPT (+8-shot)
& 62.93 & - & 59.78 & -  & 64.88 & -  \\

12 & GPT-4 (+8-shot)
& 61.89 & - & 67.44 & -  & 60.83 & -  \\


\hline \hline

\multicolumn{8}{c}{\textit{Continual pre-training LLM and then fine-tuning on supervised datasets}} \\ \hline

13 & GLM-2B
& 91.16 & +0.00 & 91.95 & +0.00 & 91.04 & +0.00 \\
14 & \quad + DKE
& 92.28 & +1.12 & 92.99 & +1.04 & 92.15 & +1.11 \\
15 & \quad + ICP
& 92.72 & +1.56 & 93.40 & +1.45 & 92.58 & +1.54 \\
16 & \quad + RCD
& 91.59 & +0.43 & 92.36 & +0.41 & 91.46 & +0.42 \\
17 & \quad + DKE + ICP 
& 93.33 & +2.17 & 93.93 & +1.98 & 93.23  & +2.19 \\
18 & \quad + DKE + ICP + RCD (a.k.a. \textbf{CPRM})
& \underline{93.64} & +2.48 & \textbf{94.42} & +2.47 & \underline{93.49} & +2.45  \\
\bottomrule[1.5pt]
\end{tabular}
}
\caption{Performance of different baselines and various continual pre-training models on the relevance task. \textbf{Bold} and \underline{underline} represent the best and second best result respectively. Improvements over variants are statistically significant with p < 0.05.} \label{tbl:performance-comp}
\end{table*}
\subsection{Baselines}
We compare our proposed CPRM model with the following baselines:
\begin{itemize}[leftmargin=*]
\item \textbf{DSSM}~\cite{shen2014learning} is a classic two-tower structure text matching model that constructs representations for the query and item independently, using cosine similarity to measure their relevance. 
\item \textbf{ReprBERT}~\cite{yao2022reprbert} is a representation-based BERT model that utilizes novel interaction strategies to balance performance and latency. 
\item \textbf{BERT}~\cite{devlin2018bert} has achieved great success on NLP tasks as an interaction-based model. Here, we concatenate the query and item as the model input for relevance modeling. 
\item \textbf{GLM}~\citep{du2022glm} is a powerful LLM architecture with various parameter sizes to suit different business scenarios. Our LLM online system is developed based on the GLM, thus all our experiments are mainly conducted on the GLM. 
\item \textbf{Qwen2}~\citep{yang2024qwen2technicalreport} is currently one of the newest and the state-of-the-art (SOTA) open-source LLMs for Chinese NLP tasks. 
\item \textbf{ChatGPT\footnote{The version is GPT-3.5-Turbo.} \& GPT-4}~\citep{openai2024gpt4technicalreport} are the SOTA closed-source LLMs. We employ the direct generation approach for relevance task evaluation. 
\end{itemize}

\subsection{Offline Experimental Results}
\textbf{Performance Comparison.}\quad Table~\ref{tbl:performance-comp} presents the performance of different baselines and various continual pre-training models on the relevance task. From the experimental results, GLM demonstrates strong competitiveness, achieving superior performance even with similar parameter numbers compared to BERT-Large (line 7 vs. line 4). We also conducts experiments on the GLM of various parameter sizes, and the results show that as the model size increases, its performance gradually improves. However, with further increases in model size, the performance gains become progressively smaller. Specifically, GLM-10B achieves only a 0.21\% improvement in AUC over GLM-5B (line 10 vs. line 9). Compared to other latest LLMs such as Qwen2, our GLM also shows impressive performance at a similar parameter scale (line 7 vs. line 5, line 8 vs. line 6). ChatGPT \& GPT-4 performed poorly compared to other SFT-based baseline systems; this is because the task data belongs to a proprietary domain, and models without SFT operations have relatively poor discriminative ability.
We conducts continued pre-training experiments on GLM-2B, and the experimental results demonstrate that all three different methods result in performance improvements compared to the baseline model. Notably, the DKE and ICP methods achieve significant performance enhancements, with respective gains of 1.11\% and 1.54\% in AUC. This is because both methods are constructed based on domain-specific data and jointly training semantically related queries or items can further enhance model performance. The experimental results also indicate that integrating different continued pre-training methods can further strength model performance (line 17 and line 18), with the combination of all three methods leading to the greatest performance gain, making it comparable to that of GLM-10B (line 18 vs. line 10). Our CPRM model achieves the highest F1 score (94.42\%) among all models compared. 
\begin{figure}[!t]
    \centering
    \includegraphics[width=0.93\linewidth]{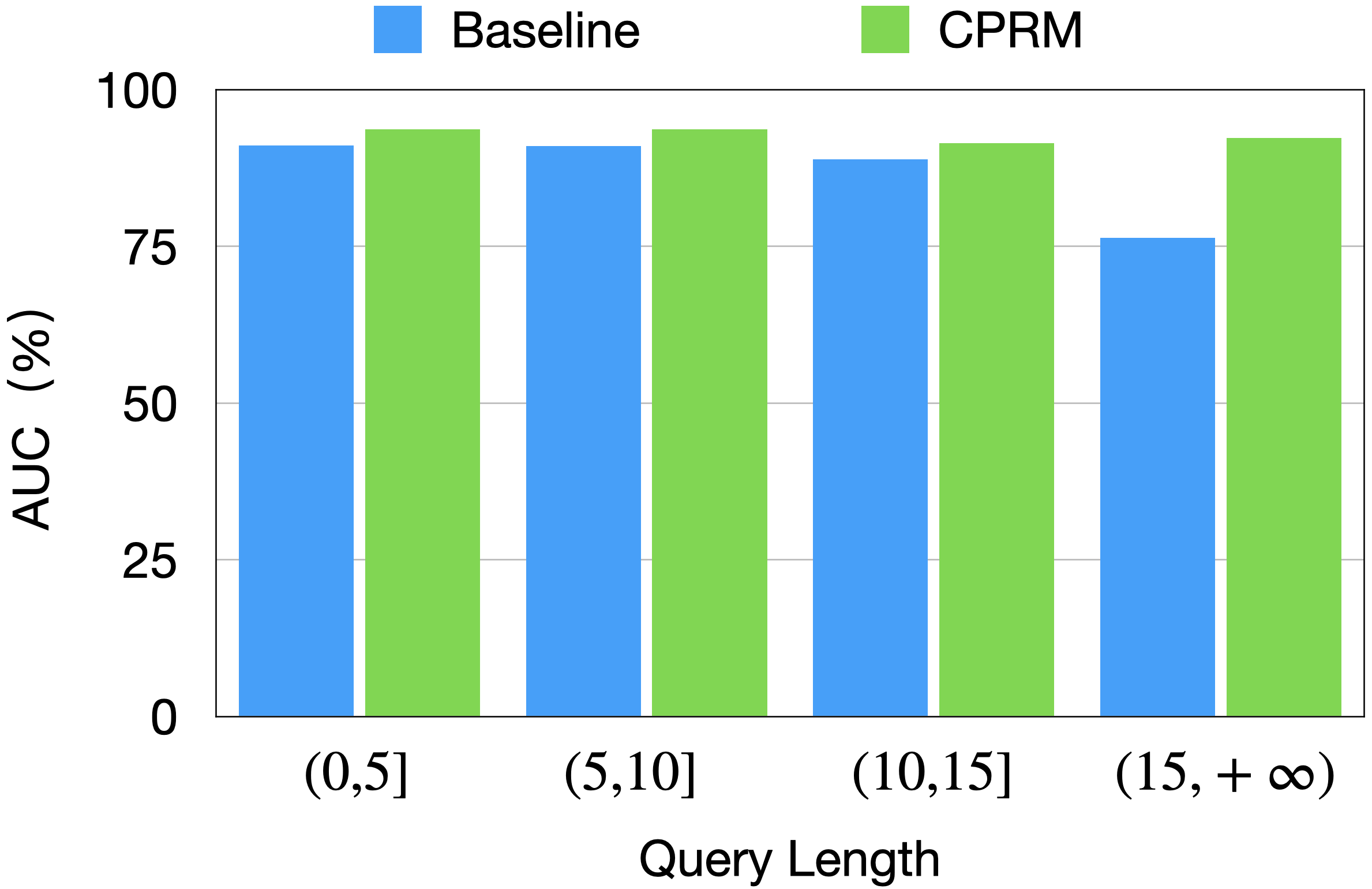}
    \caption{Performance of different query lengths.}
    \label{fig:different_length}
\end{figure}

\textbf{Analysis on Query Length.}\quad We compare the performance of different models at various query lengths on test set. As shown in Figure~\ref{fig:different_length}, our CPRM model outperforms the baseline model across all length intervals, especially on longer queries (when the length greater than 15), where the CPRM model demonstrates a significant improvement performance gains, surpassing the baseline model by 15.85\% in AUC (92.27\% vs. 76.42\%). This suggests that the CPRM model possesses a superior ability to understand and deal with long queries. We speculate that this advantage may be attributable to the ICP and RCD methods. Since the ICP method semantically aggregates historical search queries, allowing the model have the possibility encountered related long queries and to understand their semantics in the in-context pre-training process. On the other hand, the RCD method generates diverse queries, thereby enriching the model's understanding of various long query types. 

\begin{figure}[!t]
    \centering
    \includegraphics[width=0.93\linewidth]{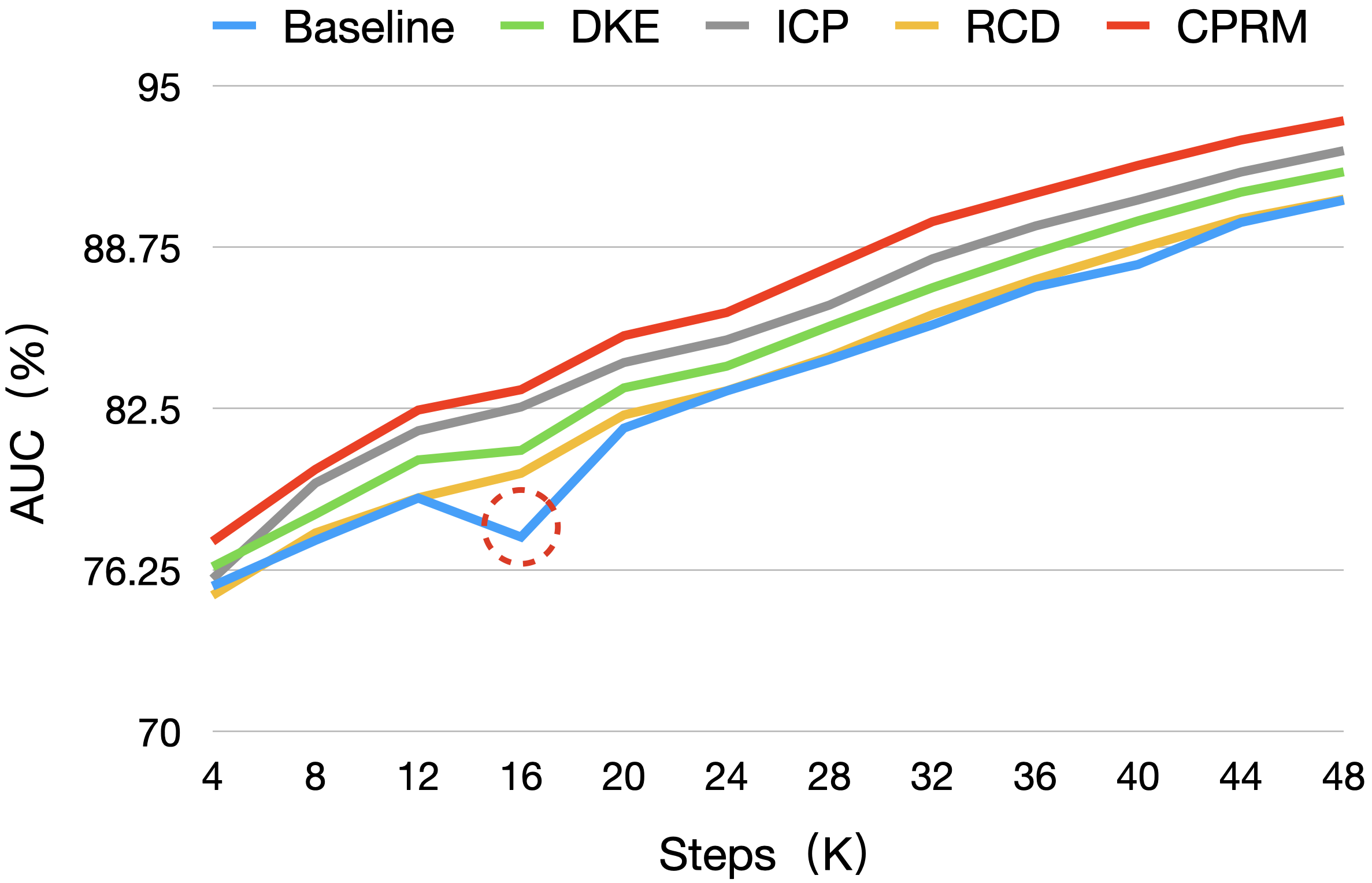}
    \caption{Performance of different training steps.}
    \label{fig:steps}
\end{figure}

\textbf{Impact of Training Steps.}\quad As shown Figure~\ref{fig:steps}, we compare models' performance with different pre-training methods across various training steps. The experimental results show that models trained with all three different pre-training methods surpasse the baseline across various training steps. The CPRM model, which combines all three methods, achieves the best performance at each step. These evidence highlights the robustness of our proposed approach. Interestingly, an phenomenon observed from the figure is that the baseline model's performance significantly decreases at the 16K training step before it gradually increases thereafter. The reason is due to the significant difference between the current task data and the data previously seen by the LLM, resulting in challenges for the LLM in fitting this domain-specific data. None of the other pre-training methods exhibits this phenomenon; instead, the performance of these models steadily improved at each training step. This indicates that our proposed methods are beneficial for the domain adaptation of LLM.
\section{Online A/B Testing}
We deploy the proposed model on the online search platform to provide search services for mini-apps, and conduct a two-week online A/B testing with 5\% proportion of the experiment traffic. 
The experimental results show that, compared to the baseline system (GLM-2B), our CPRM method yields a statistically significant increase of 0.32\% in valid PVCTR\footnote{Page view click-through rate, the number of valid clicks divided by the number of searches.} at a 95\% confidence level.
Human evaluation indicates a 0.75\% reduction in Badcase@10 metric and a 4.71\% decrease in the Error Filtering Rate\footnote{The number of relevant items that are incorrectly filtered out divided by the total number of filtered items.}. The model has now been serving search functions to mini-apps for over nine months. These results suggest that our proposed method can effectively enhance relevance models' performance in real-world search systems.

\begin{figure}[!t]
    \centering
    \includegraphics[width=1.0\linewidth]{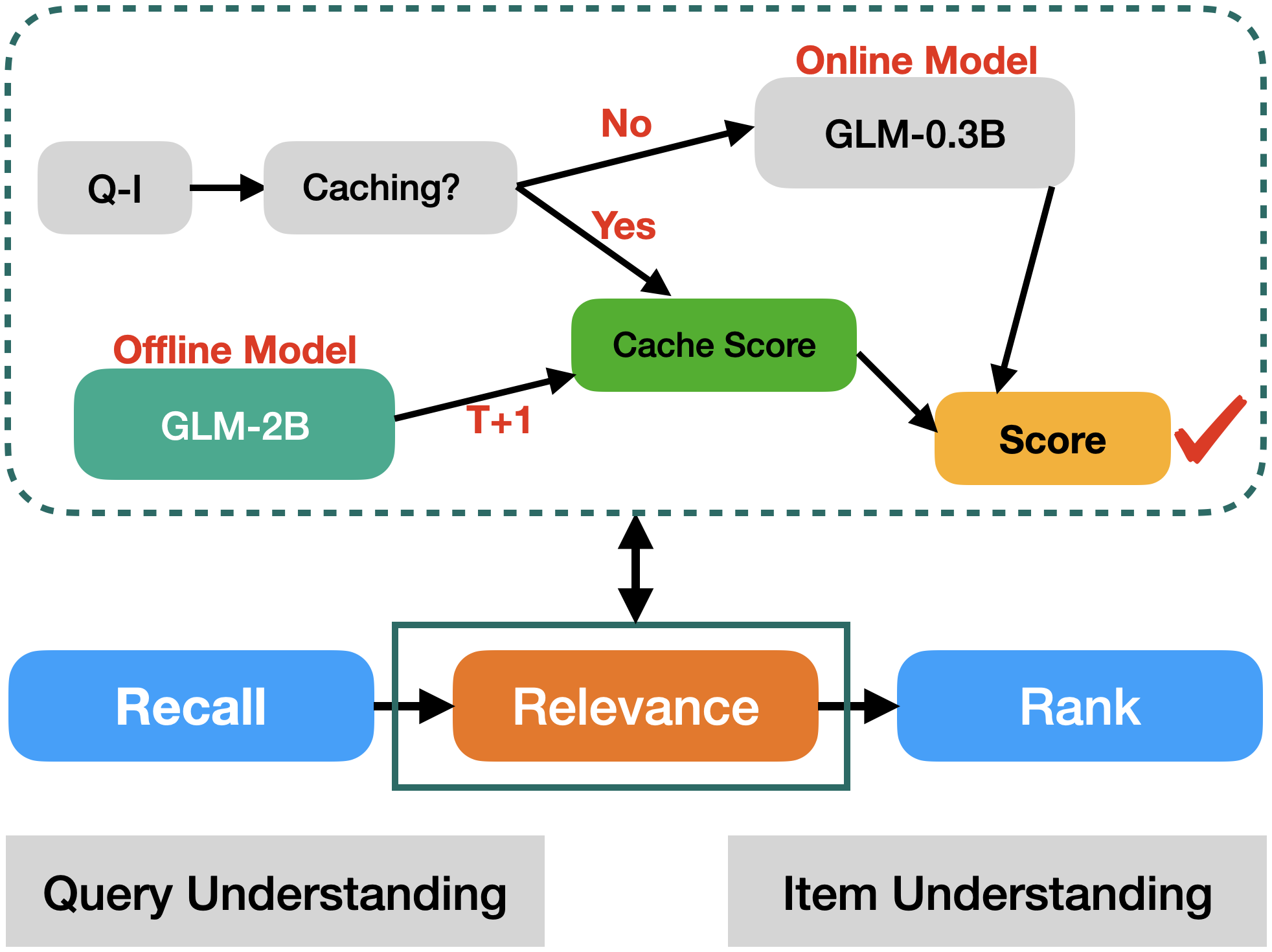}
    \caption{Deployment of the CPRM relevance model.}
    \label{fig:deploy}
\end{figure}
\section{Model Deployment}
LLMs have achieved significant performance improvements in relevance tasks, but their large parameter size leads to low inference efficiency, thus affecting their deployment online. We have designed a solution that allows for the real-time use of LLMs' relevance scores. As shown in Figure~\ref{fig:deploy}, the online relevance model for search consists of two parts: the GLM-0.3B model serves as the online model to respond to search queries in real-time, while the GLM-2B model employ a T+1 update strategy to score historical Q-I pairs and cache them offline. The online relevance service gives priority to using the cached scores from GLM-2B; if the cache does not exist, it calls on the GLM-0.3B online model. Currently, using GLM-2B's offline caching scoring has covered over 60\% of mini-app search requests, significantly alleviating the request pressure on the online model.

\section{Conclusion}
In this paper, we have investigated CPRM framework, a continued pre-training approach of LLMs tailored to relevance modeling tasks, which comprises three methods: DKE, ICP and RCD. 
Both offline experiments and online A/B testing results demonstrate that our proposed method boosts the search relevance of LLMs effectively. Our model has been successfully deployed online search platform. 


\bibliography{anthology,custom}
\bibliographystyle{acl_natbib}

\appendix
\section{Case Study} 
\begin{figure*}[!t]
    \centering
    \includegraphics[width=1.0\linewidth]{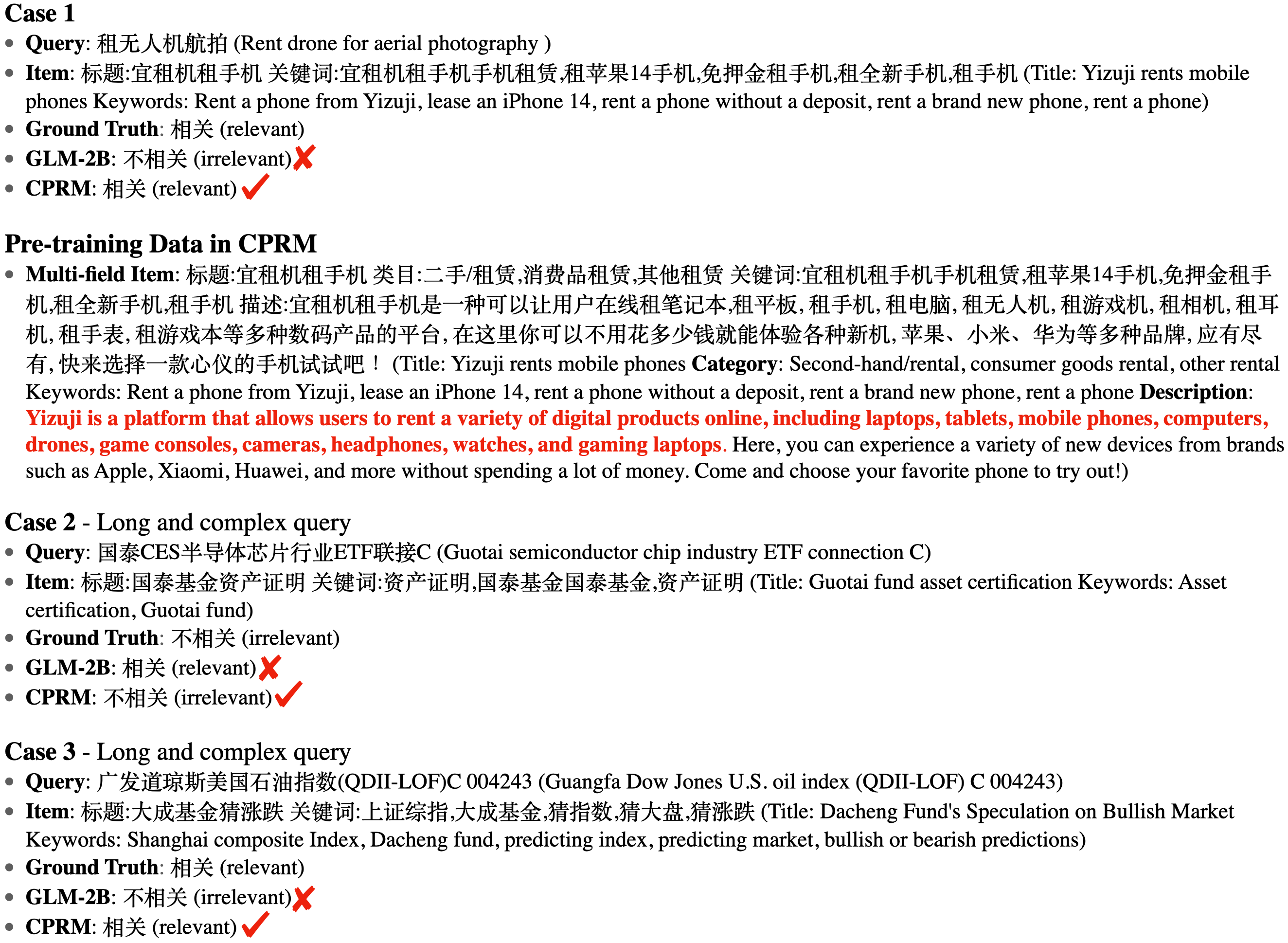}
    \caption{Case study.}
    \label{fig:case_study}
\end{figure*}

\begin{figure*}[!t]
    \centering
    \includegraphics[width=1.0\linewidth]{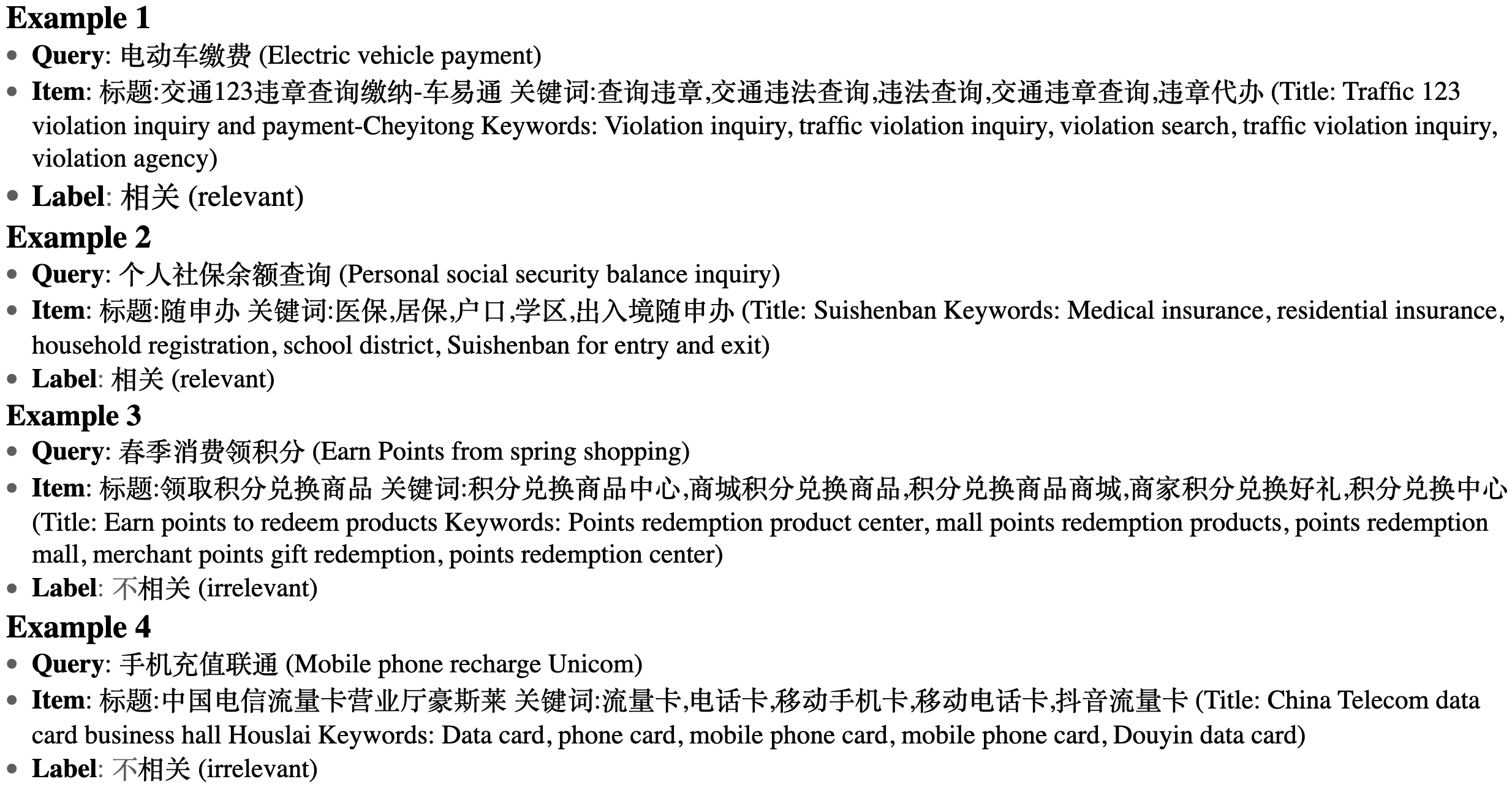}
    \caption{SFT data examples.}
    \label{fig:sft_data}
\end{figure*}

As shown in Figure ~\ref{fig:case_study}, we provide several cases to compare the relevance output results between the baseline (GLM-2B) and CPRM models. From these cases, we can observe that the CPRM method is able to supplement additional domain knowledge to correct erroneous prediction results. Furthermore, CPRM demonstrates a stronger understanding of long and complex queries (query length greater than 15) compared to the baseline. The SFT data format can be referred to in Figure ~\ref{fig:sft_data}.

\end{document}